# Pseudo-Data based Self-Supervised Federated Learning for Classification of Histopathological Images

Yuanming Zhang, Zheng Li, Xiangmin Han, Saisai Ding, Jun Wang, *Member*, *IEEE*, Shihui Ying, *Member*, *IEEE*, Jun Shi, *Member*, *IEEE*

*Abstract*—Computer-aided diagnosis (CAD) can help pathologists improve diagnostic accuracy together with consistency and repeatability for cancers. However, the CAD models trained with the histopathological images only from a single center (hospital) generally suffer from the generalization problem due to the straining inconsistencies among different centers. In this work, we propose a pseudo-data based self-supervised federated learning (FL) framework, named SSL-FT-BT, to improve both the diagnostic accuracy and generalization of CAD models. Specifically, the pseudo histopathological images are generated from each center, which contain both inherent and specific properties corresponding to the real images in this center, but do not include the privacy information. These pseudo images are then shared in the central server for self-supervised learning (SSL). A multi-task SSL is then designed to fully learn both the center-specific information and common inherent representation according to the data characteristics. Moreover, a novel Barlow Twins based FL (FL-BT) algorithm is proposed to improve the local training for the CAD model in each center by conducting contrastive learning, which benefits the optimization of the global model in the FL procedure. The experimental results on three public histopathological image datasets indicate the effectiveness of the proposed SSL-FL-BT on both diagnostic accuracy and generalization.

## I. INTRODUCTION

Cancers seriously threaten human health. Histopathological diagnosis is the "gold standard" for the diagnosis of cancers in clinical practice [1]. However, it generally suffers from the issues of low efficiency, consistency and repeatability [2]. To this end, computer-aided diagnosis (CAD) for histopathological images has attracted considerable attention in recent years [3][4]. As a classical deep learning method, convolutional neural network (CNN) and its variants have proved their effectiveness as the backbone for the CAD models of histopathological images [3][4][5][6][7][8].

It is worth noting that even the most common and accessible type of stain, such as hematoxylin and eosin (H&E), will still produce different color intensities depending on the brand, storage time, and temperature [1]. It then results in inconsistencies in the stained histopathological images among different hospitals [9]. If the training samples are only acquired from one hospital, the generalization of a CAD model then will be degraded. To this end, a potential solution is to train the CAD model with the histopathological images from multiple hospitals (*i.e.*, multi-centers). Some pioneering works have indicated the feasibility and effectiveness of multi-center learning for improving the generalization of CNN models [10]. Moreover, this manner also can alleviate the problem of small sample size (SSS), which is a common issue in the field of CAD [4].

For multi-center learning, it is a popular way to gather data from all centers together to train a model [11][12]. However, this training strategy suffers from the issues of privacy protection, data security and data ownership [13]. Federated learning (FL) then emerges as a promising solution, which can jointly train the CAD models by sharing parameters of distributed local models instead of the local data in the conventional multi-center learning paradigm [14][15]. This new multi-center learning paradigm has gained considerable attention in the field of healthcare [13][16], and it has been successfully applied for the CAD tasks [17][18], including for histopathological images [19]. However, it still cannot guarantee that the distributed CAD models fully capture the specific properties of different centers' data, because FL only shares the model parameters instead of data, and the distributed local models do not contain enough specific information.

In recent years, image synthesis has achieved remarkable performances due to the fast development of generative adversarial network (GAN) and its variants [20][21][22]. Some works have adopted the synthesized images for data augmentation to train CNN models [23][24]. Thus, if some pseudo histopathological images are generated in each center, they can contain inherent and specific properties corresponding to the real histopathological images of the center, but do not include the privacy information. Thus, it is a feasible way to share these pseudo data to pre-train the backbone of CNN model in the central server, and then further conduct FL. This strategy can promote the CAD model to learn more specific properties of each center's data and further improve the

This work is supported by National Natural Science Foundation of China (81830058, 11971296) and the 111 Project (D20031). (Corresponding authors: Jun Shi)

J. Shi, Y. Zhang, Z. Li, X. Han, S. Ding and J. Wang are with the Key Laboratory of Specialty Fiber Optics and Optical Access Networks, Joint International Research Laboratory of Specialty Fiber Optics and Advanced Communication, Shanghai Institute for Advanced Communication and Data Science, School of Communication and Information Engineering, Shanghai University, China. (Email: junshi@shu.edu.cn)

S. Ying is with the Department of Mathematics, School of Science, Shanghai University, China.

generalization ability. However, the pseudo histopathological images do not have corresponding labels for cancers, and therefore, they cannot be directly used in the same classification task as the real images to pre-train the backbone network of a CAD model.

Self-supervised learning (SSL) then provides a feasible way to explore and learn inherent information from these pseudo histopathological images, because it generates supervision directly from the training samples themselves to design pretext tasks [25][26]. SSL can effectively improve the feature representation of a backbone network for the downstream task, and it has been successfully applied to various tasks in the field of medical image analysis [27][28]. Consequently, we can develop a multi-task SSL-based FL framework to make full use of the pseudo histopathological images. In particular, an image restoration task is applied to learn the common properties of all multi-center pseudo data stored in the central server. Moreover, since we know which center a pseudo image is generated from, it can be used as label information. Consequently, we specifically design a center classification task that discriminates the source of an image generated from. This SSL task can make the pre-trained backbone learn more individual data representation of different centers to improve model generalization. Overall, the abovementioned tasks perform simultaneously and fully learn both inherently common representation across multiple centers and center-specific knowledge.

It is worth noting that although the proposed multi-task SSL with shared pseudo images under FL framework can effectively improve the generalization of CAD models, the FL still suffers from the issue of data heterogeneity due to the stain difference in different centers. In the conventional FL algorithms, the data heterogeneity will result in the drift of local models during training procedure, which then makes the objective functions of local models far from that of global model. To this end, Model-Contrastive Federated Learning (MOON) has been proposed recently, which innovatively introduces contrastive SSL into FL for model-level contrast [29]. MOON adopts the similarity between model representations to correct the local training of individual centers, and it has achieved superior performance in handling the heterogeneity of local data distribution [29]. Although MOON has the potential to alleviate the heterogeneity of histopathological images, when MOON maximizes the representation agreement between the local and global models, the contrastive operation is still inefficient due to the requirement of negative samples similar to SimCLR [30]. In fact, insufficient negative sample pairs in contrastive SSL will result in insufficient clustering, and cannot distinguish the sample difference between groups [31]. On the contrary, excessive negative sample pairs may lead to over-clustering, which makes the model difficult to learn common features for samples of the same class [31].

As a new competitive contrastive SSL algorithm, Barlow Twins (BT) proposes a contrastive objective function based on the cross-correlation matrix by minimizing the redundancy between the components of the output vectors [32], and therefore, it eliminates the redundant information expression in the representation vector as much as possible. Compared to the contrastive operation in MOON, BT has the advantage of training without negative samples, and avoids the abovementioned problems. In addition, it is more robust to the training batch size and avoids other complex implementations, such as asymmetric mechanisms and momentum encoders [33]. On the other hand, the previous contrastive learning algorithms, such as MoCo and SimCLR, build the similarity matrix in the batch dimension, while BT performs it in the feature dimension to learn a feature representation with more information, since the dimension of each feature has an independent meaning [32]. Therefore, BT has the potential to be integrated into FL to conduct model-level contrast for improving local training of individual centers.

In this work, a novel SSL-based FL (SSL-FL) framework is proposed to improve the performance of a CAD model for histopathological images. Specifically, the pseudo histopathological images are firstly generated in each center, which are fed to a specially designed multi-task SSL model to pre-train the backbone as the initial global model for further FL. The BT-based FL (FL-BT) algorithm is then developed to further effectively train the CAD models for each center with improved performance. The experimental results on four public histopathological image datasets indicate the effectiveness of the proposed SSL-based FL-BT (SSL-FL-BT).

The main contributions of this work are three-fold as follows:
1) A novel SSL-based FL framework is proposed to improve both the diagnostic accuracy and generalization of a CAD model for histopathological images. Different from the conventional FL paradigm that only shares the parameters of local models in multi-center learning, we propose to generate pseudo histopathological images from each center, which are then shared for pre-training backbone network. Thus, the specially designed SSL can capture and learn both the inherent and specific properties of data from different centers to benefit the generalization of the CAD model.
2) A multi-task SSL is further developed driven by the properties of pseudo histopathological images for pre-training backbone network of a CAD model. Specifically, the center classification task is designed to discriminate which center a pseudo image is generated from, and the image restoration task is applied to learn the common information across all the centers. This strategy helps to capture both the specific and common inherent information from multi-center pseudo histopathological images.
3) A new FL-BT algorithm is proposed to improve the performance of the global CAD model, in which BT is innovatively integrated into the FL framework, and does not works for SSL. In particular, FL-BT compares the representations learned by different models instead of different images in SSL, and it integrates a cross-correlation matrix-based contrastive objective into FL to conduct model-level contrastive learning. FL-BT minimizes the representation gaps between the local and global models to correct the local training, and therefore, it can alleviate the issue of data heterogeneity.

## II. RELATED WORK

### A. SSL for CAD of Histopathological Images

Over the last years, the fast development of deep learning has made breakthroughs in the field of CAD for histopathological images [3]. According to the size of histopathological images, the current works are developed for the whole-slide images (WSI) and patches from WSIs [4], respectively. Although lots of deep learning algorithms have been proposed in this field [34], they should be further improved due to the complexity of histopathological images and a variety of cancers.

Since it is time-consuming to annotate a large number of histopathological images for CAD, SSL is a promising approach to alleviate this problem by pre-training the model under the supervision of the data itself. For example, Hu *et al.* proposed a unified generative adversarial network to learn robust cell-level representation for classification of histopathological images [35]; Stack *et al.* applied the contrastive predictive coding to histopathology datasets, indicating that the low-level features were more effective for tumor classification [36]; Ciga *et al.* utilized SimCLR to pre-train the model on multiple histopathological datasets, which improved the performances on different downstream CAD tasks [37]. All these works demonstrate the effectiveness of SSL for CAD with limited histopathological images.

It is worth noting that the application of SSL should not only retain the center-specific information, but also mine more inherent common features from the data of all centers for FL in our task. However, the single pretext task generally cannot well explore this information. To this end, the multi-task SSL has the potential to learn more comprehensive features from training samples. In the pioneering work, Koohbanani *et al*. proposed a multi-task SSL algorithm Self-Path for histopathological images, which included three pathology-specific tasks, *i.e.*, magnification prediction, magnification Jigsaw puzzle and Hematoxylin channel prediction, to improve the model performance with limited annotations [38]. Since Self-Path can achieve superior performance over the single-task based approaches, we will also specifically design a multi-task SSL according to the data characteristics of multi-center histopathological images.

### B. Federated Learning

FL is an emerging distributed learning method, which aims to share the local model parameters in a parallel manner instead of the conventional local data[13]. FedAvg is the first FL algorithm that aggregates the local models by averaging the model weights [39]. Thereafter, FL has been successfully applied to many fields [14], such as financial, smart retail and healthcare, due to the advantages of both privacy-preserving and distributed optimization.

Recently, some variants of FedAvg have been proposed, which mainly include the following two methods: 1) Local training method, such as FedProx [40], SCAFFOLD [41] and MOON [29]; 2) Aggregation method, such as FedNova [42], FedMA [43], FedAvgM [44] and Auto-FedAvg [45]. FedProx introduced a proximal term in local training, which was calculated based on the Euclidean norm between the output of both current global model and local model [40]. FedBN was proposed to locally keep batch normalization parameters in order to mitigate feature variation in non-IID data [46]. MOON developed a model-level contrast learning strategy, whose key idea was to use the similarity between model outputs to rectify individual local training [29]. All these works show the effectiveness of FL for multiple center learning.

FL is particularly attractive for CAD now [19]. It can not only improve the generalization of CAD models, but also alleviate the SSS issue by collecting multi-center data with privacy protection. Some pioneering works have been conducted. For example, Li *et al.* proposed an FL algorithm for diagnosing the autism spectrum disorders with multi-site fMRI data, in which decentralized iterative optimization and randomization mechanism were used [17]; Andreux *et al.* introduced a local statistical batch normalization layer in the model architecture of FL, which was applied to the diagnosis of breast tumor with multi-centric histopathology datasets [47]; Yang *et al.* proposed an FL algorithm using partial networks for COVID-19 diagnosis with multiple X-ray datasets [18]; Adnan *et al.* applied the FL framework to WSIs on the data from TCGA, and they adopted the multiple instance learning (MIL) method for classification of WSIs by extracting multiple patches from the WSIs [48]. These works indicate that FL can effectively improve the model performance in local servers together with privacy-preserving.

However, the existing FL methods cannot sufficiently handle the gap between the local models and the central model, resulting in limited learning performance.

### C. GAN in Histopathological Images

Due to the success of GAN in computer vision, it has also been widely used in different medical image tasks, such as image reconstruction, image segmentation and lesion detection [49][50][51]. Moreover, they can help alleviate the problems of small sample size and limited annotation in medical imaging applications. For example, Madani *et al.* applied the GAN-based data augmentation for the CAD model of Chest X-ray, and achieved superior performance over the traditional augmentation strategy [23]; Adar *et al*. used conditional GANs to generate synthetic CT images to improve the performance of liver lesion classification [24]; Elmas *et al.* proposed a FL-based MRI reconstruction algorithm, which implemented cross-cite learning with generative MRI prior and the following prior adaptation to improve reconstruction performance [49]. These works demonstrate the effectiveness of GAN in different medical image processing tasks.

Some works have applied GAN to the field of histopathological image processing, such as color normalization, image enhancement and data augmentation [52][53][54]. For example, CycleGAN was effectively used for color normalization of breast histopathological images, which then eliminated the selection of representative reference slides by pathologists [52]; SRGAN was applied to simultaneously increase image resolution and reduce image noise for breast histopathological images [53]; cGAN was adopted to synthesize realistic cervical histopathology images for augmenting the training dataset so as to improve the performance of trained model [54]. All these works indicate the emerging applications of GAN for histopathological images.

Different from these previous works, we propose to adopt GAN to generate pseudo histopathological images in different centers, and then share these images in the central server for training the global model under the FL framework. This idea breaks through the limitation of traditional FL that only shares model parameters. The specially designed SSL can capture and learn both the inherent and specific properties of data based on the shared pseudo images to further improve the generalization of the global CAD model.

## III. METHODOLOGY

Fig. 1 shows the overall pipeline of the proposed SSL-FL-BT, which includes two stages, *i.e.*, SSL stage and FL stage. In the SSL stage, the pseudo histopathological images are firstly generated in each center with a GAN. The specially designed multi-task SSL is then performed on all the pseudo images to pre-train the backbone network. Here, the center classification and image restoration tasks are designed as the multiple pretext tasks, and both tasks share the backbone. The pre-trained backbone is then used as the initialization network in the subsequent FL stage, and it is trained by the proposed FL-BT with multi-center real histopathological images. In the testing stage, a histopathological image is fed to the corresponding CAD model in a center for cancer diagnosis.

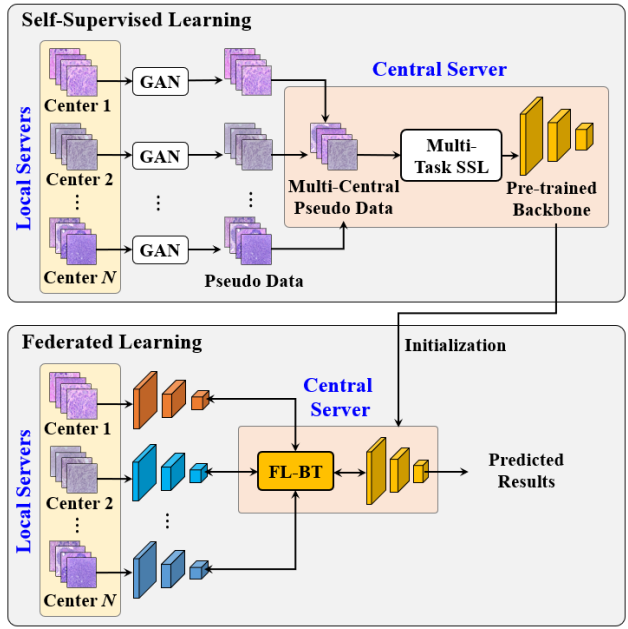

Fig. 1. The pipeline of the proposed SSL-FL-BT, which includes two stages, *i.e.*, SSL stage and FL stage. In the SSL stage, the specially designed multi-task SSL is performed on all the pseudo images to pre-train the backbone network. In the FL stage, the pre-trained backbone is used as the initialization network for the proposed FL-BT model.

### A. Multi-task SSL for FL

Since the stained histopathological images have inconsistencies among different centers, we propose to share the pseudo histopathological images without privacy information for FL, which can provide more heterogeneous center-specific information of each center for the CAD model, and further improve its generalization. Here, we specifically design a multi-task SSL to capture and learn both the center-specific information and common inherent representation according to the data characteristics of multi-center pseudo histopathological images. The overall pipeline of our proposed multi-task SSL is shown in Fig. 2.

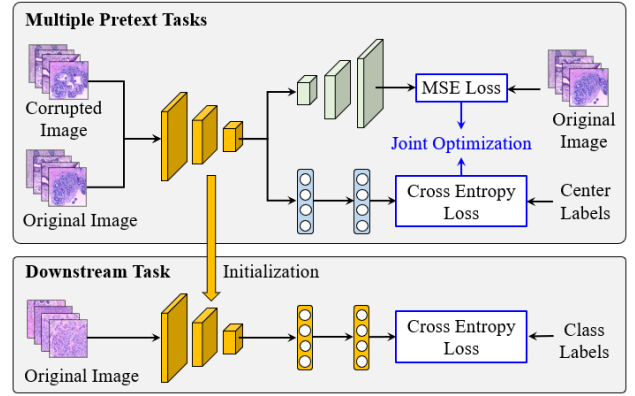

Fig. 2. The pipeline of the proposed multi-task SSL. The pseudo data are shared in the central server to pre-train the backbone network by multi-task SSL. Two pretext tasks are designed, *i.e.*, the center classification task and image restoration task.

As shown in Fig. 2, the pseudo data are firstly generated in each center through the GAN model, which are then shared to the server for pre-training backbone network by SSL. Two pretext tasks are then designed, *i.e.*, the center classification task and image restoration task. The former pretext task predicts which center the synthetic data belong to. It can explore more specific properties of data in each center. While the latter pretext task restores the corrupted images to their original pseudo images, which can learn more inherent information of the data collected from different centers.

**Pseudo Image Generation**: In order to generate high-fidelity pseudo histopathological images, the multi-scale gradient generative adversarial network (MSG-GAN) algorithm is adopted in this work, which provides high-quality synthesized images for the following multi-task SSL [55]. In particular, each center individually trains an MSG-GAN.

MSG-GAN introduces a multiscale gradient technique that allows the gradients flow to propagate from the discriminator to the generator at multiple scales. This technique improves the stability of training for image synthesis on data with different sizes, resolutions and domains. Compared to other GANs and their variants, MSG-GAN can boost the performance in most of cases. The detailed information about the MSG-GAN can be referred to [55].

The quality of the generated images is evaluated by frechet inception distance (FID) score in this work, which is a measure to calculate the distance between the feature vectors of real images and generated images [56]. The pseudo histopathological images with small FID score will be selected for the following SSL. The real histopathological images and generated pseudo examples are shown in Fig. 3.

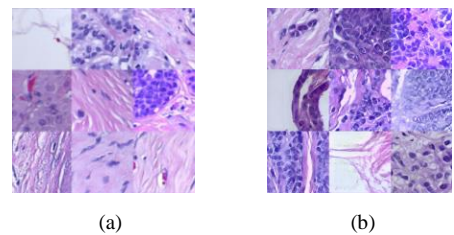

(a)           (b)

Fig. 3. (a) Real histopathological images and their corresponding (b) Pseudo histopathological images, where the images of three rows in (a) are acquired for Center 1, Center 2 and Center 3, respectively.

**Multiple Pretext Tasks:** Two SSL pretext tasks are developed based on the characteristics of pseudo histopathological images for pre-training backbone, *i.e.*, the center classification task and image restoration task.

The center classification task tries to predict which center a pseudo image is generated from, and thus the center identity document (ID) is considered as the label. Since the pseudo images are generated based on each center's data, these images contain center-specific information extracted from the real histopathological images of the corresponding center. Thus, this pretext task can effectively learn the heterogeneous characteristics pseudo images generated from each center.

The SSL image restoration task is applied to learn detailed context information and high-level representation in histopathological images, which contain the common characteristics across centers. Specifically, we randomly swap patches in the original pseudo images to generate the corrupted ones [27]. These corrupted images are then fed to the backbone network to restore the original pseudo images as ground truths.

To conduct two pretext tasks in a unified framework with a single shared network, the hard parameter sharing is utilized to construct a multi-task learning architecture [57]. In our implementation, the commonly used ResNet50 is used as the shared backbone, followed by a classification branch and a reconstruction head [58].

For the center classification task, the cross-entropy (CE) loss $L_{CE}$ is utilized for the SSL classification task, which can be given as follows:

$$L_{CE} = -\frac{1}{M}\sum_k \sum_{n=1}^N a_{kn}\log(u_{kn}) \quad (1)$$

where $a_{kn} \in \{0,1\}$ is an indicator, which takes value 1 if and only if the label of $k$-th sample is $n$, $u_{kn}$ denotes the probability of the $k$-th sample coming from the $n$-th center, and $M$ denotes the number of pseudo images.

For the image restoration task, the mean squared error (MSE) is adopted as the objective function for the SSL image restoration task. Given a corrupted image $Q_k$ and the reconstruction sub-network $G(\cdot)$, the MSE loss is formulated as:

$$L_{MSE} = \frac{1}{M}\sum_{k=1}^M \|G(Q_k) - V_k\|^2 \quad (2)$$

where $G(Q_k)$ and $V_k$ denote the restored image and the corresponding ground truth, respectively.

The overall loss $L_{SSL}$ for our multi-task SSL is then formulated as:

$$L_{SSL} = L_{CE} + L_{MSE} \quad (3)$$

The pre-trained backbone is then used as the initialization for the followed FL stage. This multi-task SSL strategy can capture both the specific and common inherent information from the multi-center pseudo histopathological images.

### B. FL-BT for Histopathological Images

The generalization of a CAD model for histopathological images is generally limited by the training samples only from a single center, because the stained images have different data distributions in different hospitals. FL can improve both the diagnostic accuracy and generalization ability of CAD models with multi-center histopathological images, while private information can also be protected [59].

The existing FL methods cannot well handle the heterogeneity of multiple local data distribution. MOON is proposed to address this issue, which adopts the similarity among model representations to correct the local training of individual centers [29]. However, the robustness of MOON should be further improved, since the trivial implementations are applied in the local training process. To this end, we propose a novel FL-BT to train the CAD model with multi-center histopathological images.

**Network Architecture of Local Model:** The architecture of each local network for FL-BT is shown in Fig. 4, which consists of a base encoder, a projector network, and an output layer. The base encoder is the widely used ResNet50, which learns the feature representation from input real histopathological images. The projector network is then adopted to map the representation to a feature space with a fixed dimension. Finally, the output layer is used to predict the classification results for each cancer class.

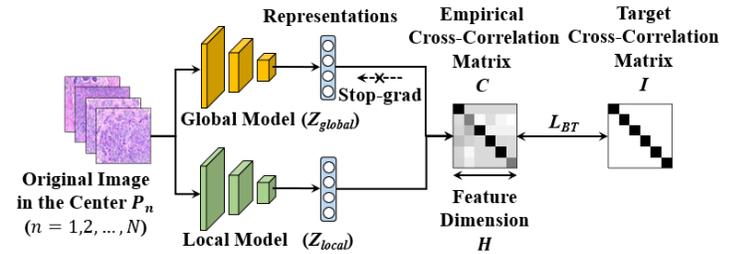

Fig. 4: Flowchart of the FL-BT algorithm. FL-BT feeds the same image into the local model and the global model, respectively, and then calculates the covariance matrix of the corresponding two features. FL-BT optimizes the statistical properties that make the cross-correlation tend to be the identity matrix.

**Local Training:** Suppose there are $N$ centers, which are denoted as $P_1$, ..., $P_N$. Center $P_n$ has a local dataset $\mathcal{D}_n$ with histopathological images $X_n$ and the corresponding label $Y_n$, $n = 1,2,...,N$. The proposed FL-BT aims to learn a global model $W$ over the local dataset $\mathcal{D}_n$ with the help of a global model in the central server.

To train the model $W_n$ in each center, the proposed FL-BT assumes that the pre-trained global model $W$ trained by multi-task SSL is set as the initial model $W_n$ in center $P_n$. The histopathological images $X_n$ are fed to the base encoders of both global model and local model to generate representations $z_{global}$ and $z_{local}$, respectively. The projector network then maps the $z_{global}$ and $z_{local}$ to the fixed feature dimension $H$. Finally, the classification result is predicted by the output layer.

We further define $F_w(\cdot)$ as the whole network, $R_w(\cdot)$ as the network before the output layer with model weight $w$, and $t$ as the $t$-th communication round. We extract the representation of $X_n$ from both the global model $W^t$ (i.e., $z_{global} = R_{w^t}(X_n)$) and the local model being updated $W_n^t$ (i.e., $z_{local} = R_{w_n^t}(X_n)$), respectively.

The local objective function contains two parts: $\mathcal{L}_{sup}$ and $\mathcal{L}_{FL\text{-}BT}$. The former part $\mathcal{L}_{sup}$ is a cross-entropy loss in the supervised learning manner, while the second part $\mathcal{L}_{FL\text{-}BT}$ is the contrastive loss in our proposed FL-BT.

Specifically, the supervision loss $\mathcal{L}_{sup}$ in FL-BT can be given as:

$$L_{sup} = CrossEntropyLoss(F_{w_n^t}(X_n), Y_n) \quad (4)$$

While the contrastive loss $\mathcal{L}_{FL\text{-}BT}$ designed in FL-BT can be formulated as:

$$L_{FL-BT} = \sum_i (1 - C_{ii})^2 + \lambda \sum_i \sum_{j \neq i} C_{ij}^2 \quad (5)$$

where $\lambda$ is the weight to trade off the importance of the first and second terms; $C$ denotes the cross-correlation matrix computed between the outputs of the two branches along the batch dimension, which is a square matrix with size the dimensionality of the network's output; and both $i$ and $j$ are the vector dimensions of the network outputs. More specifically, $C_{ij}$ can be calculated by:

$$C_{ij} = \frac{\sum_b z_{local}^{b,i} z_{global}^{b,j}}{\sqrt{\sum_b (z_{local}^{b,i})^2} \sqrt{\sum_b (z_{global}^{b,j})^2}} \quad (6)$$

where the superscript $b$ denotes batch samples. When $i = j$, we can get $C_{ii}$. The loss of FL-BT includes two parts, i.e., the invariance term and the redundancy reduction term. Among them, the invariance term makes the positive examples closer to each other in the representation space, and the redundancy reduction term decorrelates the different components of the embedding vector by making the off-diagonal elements of the cross-correlation matrix to 0. This decorrelation reduces the redundancy between outputs, make the outputs only contain non-redundant information about the samples. Therefore, it eliminates the redundant information expression in the representation vector as much as possible, making FL-BT can effectively optimize the FL procedure.

The definition of the whole loss function can be given by:

$$L_n = L_{sup}(w_n^t; (X_n, Y_n)) + \mu \mathcal{L}_{FL-BT}(w_n^t; w^t; X_n) \quad (7)$$

where $\mu$ is the factor to balance the weight of contrastive loss $\mathcal{L}_{FL\text{-}BT}$.

The local objective is to minimize:

$$\min_{w_n^t} \mathbb{E}_{(X_n,Y_n) \sim \mathcal{D}_n} [L_{sup}(w_n^t; (X_n, Y_n)) + \mu \mathcal{L}_{FL-BT}(w_n^t; w^t; X_n))] \quad (8)$$

In each round, the server sends the global model to the centers, receives the local model from the centers, and updates the global model using weighted averaging. In local training, each model uses stochastic gradient decent to update the parameters with the local data, the objective is shown in Eq. (8).

The model-contrastive loss compares representations learned by different models, and the contrastive loss compares representations of different images in FL-BT. It is worth noting that the conventional BT calculates the covariance matrix of the two representations after inputting two views of an image into the same network, while FL-BT feeds the same image into the local model and the global model to output two representations, respectively, and then calculates the covariance matrix of the two output representations.

**Global Aggregation:** After the local training in each center, the updated model parameters $w_n$, which $n = 1, ..., N$, in local models are then sent to the central server to implement the model aggregation.

FL-BT seeks to minimize the following objective function for model training:

$$\min_w L(w) = \sum_{n=1}^N \alpha_n L_n \quad (9)$$

where $N$ denotes the number of the centers, $\alpha_n$ represents the importance of the $n$-th center with $\sum_n \alpha_n = 1$.

In this work, we adopt the classical FedAvg as the aggregation method [39], in which $w_n^t$ are averaged as the global model. In communication round $t$, the updated parameters for the global model can be formulated as the follows:

$$w^{t+1} \leftarrow \sum_{n=1}^N \frac{m_n}{M} w_n^t \quad (10)$$

where $m_i$ denotes the number of images in center $P_n$, and $M$ denotes the total number of images.

Then, the updated parameters of the global model are deployed to all the local servers for the local models, which can be formulated as:

$$\forall_n w_n^t \leftarrow w_n^t - \eta g_n \quad (11)$$

where $\eta$ denotes the learning rate for model optimization and $g_i$ denotes the gradients at each local model. The final model is obtained after several communication rounds.

The detailed scheme of FL-BT is shown in Algorithm 1.

---

**Algorithm 1:** The FL-BT framework

**Input**: local datasets, number of communication rounds $T$, number of local epochs $E$, number of classes, number of centers $N$, learning rate $\eta$

**Output**: The final model $W^T$

1: **Server executes:**
2: initialize $w^0$
3: **for** $t = 0, 1, \cdots, T - 1$ **do**
4:     for $n = 1, 2, \cdots, N$ **in parallel do**
5:         send the global model $w^t$ to $P_n$
6:         $w_n^t \leftarrow$ **PartyLocalTraining**$(n, w^t)$
7:     $w^{t+1} \leftarrow \sum_{n=1}^N \frac{m_n}{M} w_n^t$
8: return $w^T$
9: **PartyLocalTraining**$(n, w^t)$:
10: $w_n^t \leftarrow w^t$
11: **for** epoch $e = 1, 2, \cdots, E$ **do**
12:     for each batch b $= \{X_n, Y_n\}$ of $D^n$ do
13:         $l_{sup} \leftarrow CrossEntropyLoss(F_{w_n^t}(X_n), Y_n)$
14:         $z \leftarrow R_{w_n^t}(X_n)$
15:         $z_{glob} \leftarrow R_{w^t}(X_n)$
16:         $\mathcal{L}_{FL-BT} \leftarrow \sum_j (1 - C_{ii})^2 + \lambda \sum_i \sum_{j \neq i} C_{ij}^2$
17:         $l_n \leftarrow l_{sup} + \mu l_{FL-BT}$
18:         $w_n^t \leftarrow w_n^t - \eta g_n$
19: **return** $w_n^t$ to server

---

**Comparisons with MOON:** MOON is a simple and effective approach for FL. Inspired by MOON, we propose FL-

BT that performs the model-level contrastive learning in FL. The FL-BT and MOON have the following differences:

1) MOON requires negative sample pairs for model-level contrastive learning, while our FL-BT does not require negative sample pairs for model contrastive learning. Therefore, FL-BT avoids under-clustering and over-clustering problems.
2) The mathematical principle of FL-BT is different from that of MOON. MOON directly optimizes the geometric properties of feature space. It pulls the positive samples closer and pushes the negative samples farther, allowing the feature space to be clustered by classes. While FL-BT optimizes statistical properties, and favors the cross-correlation towards the identity matrix rather than the geometric properties of the feature space.
3) The similarity matrix of MOON is developed based on the batch-wise, while that of FL-BT is developed based on the sample-wise. Since each sample has its own characteristics, the sample-wise learning criterion of FL-BT helps to learn superior representations for the classification.

## IV. EXPERIMENTS

### A. Datasets and Data Preprocessing

The proposed algorithm was evaluated on four public breast histopathological datasets: the 2015 Bioimaging Challenge Dataset [60], the 4th Symposium in Applied Bioimaging Dataset [61], the ICIAR 2018 Grand Challenge on Breast Cancer Histology Images Dataset [62] and the Databiox Dataset [63], which were introduced as follows:

1) The 2015 Bioimaging Challenge Dataset [60]

The 2015 Bioimaging Challenge Dataset (https://rdm.inesctec.pt/dataset/nis-2017-003) includes high-resolution (2048×1536 pixels), uncompressed, and annotated hematoxylin and eosin (H&E) stained images. All the images were digitized with the magnification of 200× and pixel size of 0.42μm×0.42μm. These images were labeled by two pathologists, and the disagreement cases between pathologists were discarded.

2) The 4th Symposium in Applied Bioimaging Dataset [61]

The 4th Symposium in Applied Bioimaging Dataset has 140 high-resolution (2048×1536 pixels) annotated HE-stained images. The images were all digitized under the same acquisition conditions with a magnification of 200x. The dataset has been assembled and annotated by two pathologists. The dataset is publicly available at http://www.bioimaging2015.ineb.up.pt/challenge_overview.html.

3) The ICIAR 2018 Grand Challenge on Breast Cancer Histology Images Dataset [62]

This dataset includes microscopy images annotated by two expert pathologists. The images with divergence between normal and benign classes were then discarded. The remaining doubtful cases were confirmed via imunohistochemical analysis. The provided images had the same size of 2048 × 1536 pixels and a pixel scale of 0.42 μm × 0.42 μm. The data is publicly available from the BACH challenge website: https://iciar2018-challenge.grand-challenge.org/.

4) The Databiox Dataset [63]

The Databiox dataset is a histopathological image dataset for grading breast invasive ductal carcinoma (IDC) into three categories: grade I, grade II, and grade III. It includes 922 images in four magnification levels, i.e. 4×, 10×, 20×, and 40×. We then selected the high-magnification 40× images (131 with grade I, 180 with grade II, and 143 with grade III) for experiments, since this subset had the largest number of samples with clearer structure and morphology about tissues. Finally, after removing the surrounding non-tissue regions, all images were then cropped into the size of 2048×1536.

The detailed information about the four datasets is given in Table I. All the histopathological images were stained by hematoxylin and eosin (H&E). Fig. 5 shows some example images from four datasets.

TABLE I
DETAILED INFORMATION ABOUT THE FOUR DATASETS

| Datasets | Classes/Numbers | Center |
|---|---|---|
| 2015 Bioimaging Challenge Dataset | Normal Tissues:64<br>Benign Lesions:78<br>In Situ Carcinomas:72<br>Invasive Carcinomas:71 | Center 1 |
| The 4th Symposium in Applied Bioimaging Dataset | Normal Tissues:30<br>Benign Lesions:30<br>In Situ Carcinomas:30<br>Invasive Carcinomas:30 | Center 2 |
| ICIAR 2018 Grand Challenge | Normal Tissues:100<br>Benign Lesions:100<br>In Situ Carcinomas:100<br>Invasive Carcinomas:100 | Center 3 |
| Databiox Dataset | Grade I: 131<br>Grade II:180<br>Grade III:143 | Additional Center |

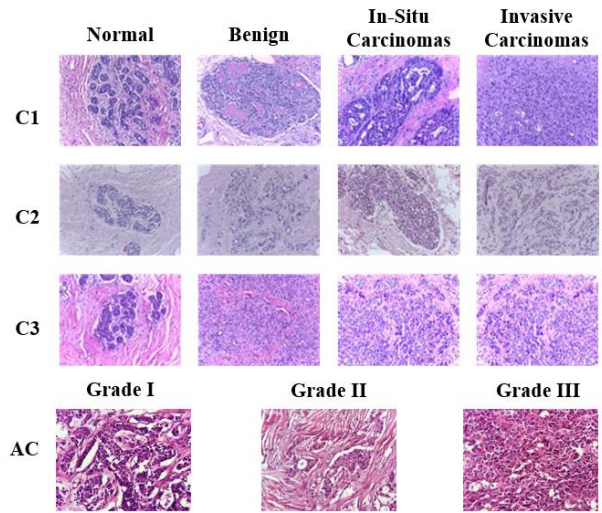

Fig. 5: The example histopathological images from four datasets.

Since the former three datasets, namely the 2015 Bioimaging Challenge Dataset, the 4th Symposium in Applied Bioimaging Dataset, the ICIAR 2018 Grand Challenge on Breast Cancer Histology Images Dataset, had the same classes, *i.e.*, normal tissues, benign lesions, in situ carcinomas, and invasive carcinomas, these datasets were then used as three centers for FL in this work, which were denoted as Center 1 (C1), Center 2 (C2) and Center 3 (C3), respectively. The Databiox dataset was adopted as an Additional Center (AC) to verify the generalization of the model, and the final global model was used as the initialization model for training.

### B. Experimental Setup

To validate the effectiveness of the proposed SSL-FL-BT, the following related algorithms were compared:
1) ResNet50 [58]: The ResNet50 was directly trained only with the histopathological images from one center, which was a single-center based CAD without FL.
2) FedAvg [39]: FedAvg was selected for comparison as a classical FL algorithm, which utilized fixed weights to average the local models for the optimization of the global model.
3) FedProx [40]: FedProx was also selected for comparison as a classical FL algorithm, which added a proximal term in the aggregation method proposed by FedAvg to stabilize the convergence.
4) FedBN [46]: FedBN was compared as a stage-of-the-art FL algorithm, which aggregated the local models without sharing parameters in BN layers to obtain the global model.
5) MOON [29]: MOON was compared as the contrastive learning-based FL algorithm, which adopted contrastive learning to reduce the gaps between the local models and the global model.

It is worth noting that all the compared algorithms adopted ResNet50 as the backbone.

On the other hand, an ablation experiment was conducted to evaluate the effectiveness of the multi-task SSL in SSL-FL by comparing SSL-FL-BT with the following variants:
1) FL-BT: FL-BT directly trained the CAD models of all centers without the pre-trained backbone by multi-task SSL.
2) SSL-C-FL-BT: This algorithm only designed the center classification task as the pretext task for SSL, and then conducted the SSL-based FL.
3) SSL-R-FL-BT: This algorithm only designed the restoration task as the pretext task for SSL, and then conducted the SSL-based FL.
4) SSL-Fed-R: It is a two-stage learning variant, which first performs SSL on the original image in each center individually in the first stage, and then the pre-trained models from different centers are used as the initialized models for the followed FL stage in the second stage. Here, we adopted image restoration task as the SSL pretext task.
5) SSL-Fed-S: It is the same variant as SSL-Fed-R, but we adopted a typical contrastive learning task, named SimCLR, instead of the image restoration task as the SSL pretext task.
6) FedSSL-R: It is also a two-stage learning variant. In the first stage, the federated self-supervised learning (FedSSL) is performed on the original image [64], which includes three key steps: 1) model local pre-training in clients; 2) pre-trained model aggregation in the server; and 3) model communication (upload and update) between the server and clients. In the second stage, these pre-trained models are used as initialized modes for the following FL, which is same as that of SSL-Fed-R. Here, we adopted image restoration task as the SSL pretext task.
7) FedSSL-S: It is the same variant as FedSSL-R, but we adopted a typical contrastive learning task, named SimCLR,

instead of the image restoration task as the SSL pretext task.

In each round of FL, the updated local models of different centers were transferred to the server to further update the global model. After training, each center adopted the global model for diagnosis. Therefore, for each algorithm, we reported the performance of the global model on the three centers as the final result.

The classification accuracy, precision, recall, and F1-score were used as evaluation indices, which were computed as follows:

$$\begin{cases} Accuracy = \frac{TP+TN}{TP+TN+FP+FN} \\ Precision = \frac{TP}{TP+FP} \\ Recall = \frac{TP}{TP+FN} \\ F1 = 2 \times \frac{Precision \times Recall}{Precision+Recall} \end{cases} \quad (12)$$

where TP is the number of true positive, TN is the number of true negative, FP is the number of false positive and FN is the number of false negative. The results were reported in the format of mean ± SD (standard deviation). The precision-recall (PR) curve and average precision (AP) were also used to evaluate the performance.

To verify the generalization of the global model, the value of global test average (GTA) was further used to quantitatively measure the generalization ability of the global model [45], which computed the average accuracy, precision, recall and F1-score values on the results of three centers, respectively.

In addition, the additional fourth dataset (Databiox Dataset) in AC was further used to verify the generalization of the global model. In particular, since the former three datasets have different disease classes from that of the fourth dataset, the global model trained through the FL paradigm under three centers was used as the pre-trained model for the last dataset. That is, the network structure before the last layer of the pre-trained model remained unchanged, and the output number of the final fully connected layer was changed from four to three. This backbone was then fine-tuned with the training set of the fourth dataset.

The five-fold cross-validation strategy was applied to all algorithms on four datasets. That is, three folds were used as the training set, one fold as the verification set, and the last fold as the testing set [65].

### C. Implementation Details

All histopathological images were resized to 256×256 for model training. MSG-GAN was used for pseudo images generation. The learning rate for the generator and discriminator was 0.003, while the number of epochs for training was 100. Each center generated 1000 images with the size of 256×256. Data augmentation was conducted on all datasets for all algorithms, including rotation (90°, 180°, 270°) and horizontally flipping.

ResNet50 was adopted as the backbone for multi-task SSL and FL. For each FL algorithm, the model was trained for 300 rounds of 1 local epoch using a batch size of 4. The weight $\mu$ for contrastive loss in Eq. (4) was set to 0.01, while the trade-off weight $\lambda$ in Eq. (7) was set to 0.005. The stochastic gradient descent (SGD) was used for the optimization of each algorithm with the learning rate of 0.001. All algorithms were implemented on Pytorch.

## V. EXPERIMENTS RESULTS

### A. Results on Multi-Center Datasets

Table II to IV show the results of different algorithms on C1, C2 and C3 datasets, respectively. It can be found that all the FL algorithms achieve superior performances to ResNet50, which is a single-center based approach, indicating that FL algorithms can effectively improve the performances of a CAD model with multi-center data. Moreover, the proposed SSL-FL-BT outperforms all the compared FL algorithms with statistical significance on all indices in all three datasets, while FL-BT also gets significantly improvements over the compared ResNet50, FedAvg, FedProx, FedBN, and MOON algorithms.

In particular, SSL-FL-BT achieves the best mean classification accuracy of 96.06±0.57%, precision of 96.21±0.46%, recall of 96.15±0.53%, and F1-score of 96.03±0.56% on C1. It improves at least 3.82%, 3.33%, 3.65%, and 3.73% on the corresponding indices compared to FedAvg, FedProx, FedBN, and MOON, suggesting that both multi-task SSL and FL-BT effectively improve the performance. On the other hand, it also can be observed that the proposed FL-BT achieves the second-best results, and gets the improvements of 1.18%, 1.16%, 1.14% and 1.17% on the corresponding indices, respectively, compared to other FL algorithms except for our proposed SSL-FL-BT, which demonstrates the effectiveness of BT in the proposed FL-BT. SSL-FL-BT also gains the best results of 96.66±1.86%, 97.14±1.60%, 96.66±1.86%, and 96.64±1.88% on the accuracy, precision, recall, and F1-score, respectively, on C2 dataset, which improves at least 4.16%, 3.03%, 4.16%, and 4.39%, on the corresponding indices, respectively, compared to other FL algorithms. The FL-BT again achieves the second-best performance by improving at least 1.67%, 1.25%, 1.67%, and 1.80% on the accuracy, precision, recall, and F1-score, respectively, over other compared FL algorithms. Moreover, SSL-FL-BT obtains the best results of 94.50±0.68%, 94.85±0.83%, 94.50±0.68%, and 94.46±0.69% on the accuracy, precision, recall, and F1-score, respectively, on C3 dataset, which improves 3.25%, 3.64%, 3.25% and 3.21% on the corresponding indices, respectively, compared to FedAvg, FedProx, FedBN, and MOON. Besides, The FL-BT is second to SSL-FL-BT and outperforms all the other compared FL algorithms. We can see that FL-BT improves of 1.25%, 1.81%, 1.25%, and 1.22% on the accuracy, precision, recall, and F1-score on C3. All these results suggest the effectiveness of our proposed SSL-FL-BT.

TABLE II
CLASSIFICATION RESULTS OF DIFFERENT ALGORITHMS ON C1 (UNIT: %)

| Algorithms | Accuracy | Precision | Recall | F1-score |
|---|---|---|---|---|
| ResNet50 | 89.26±2.33†* | 90.31±2.68†* | 89.21±2.08†* | 89.26±2.24†* |
| FedAvg | 90.83±1.50†* | 91.84±1.79†* | 90.83±1.56†* | 90.73±1.65†* |
| FedProx | 91.50±1.65†* | 92.14±1.86†* | 91.75±1.55†* | 91.58±1.78†* |
| FedBN | 91.84±1.42†* | 92.48±1.55†* | 92.08±1.41†* | 91.92±1.56†* |
| MOON | 92.24±1.70†* | 92.88±1.66†* | 92.50±1.56†* | 92.30±1.84†* |
| **FL-BT** | **93.42±1.33*** | **94.04±1.52*** | **93.64±1.22*** | **93.47±1.46*** |
| **SSL-FL-BT** | **96.06±0.57** | **96.21±0.46** | **96.15±0.53** | **96.03±0.56** |

TABLE III
CLASSIFICATION RESULTS OF DIFFERENT ALGORITHMS ON C2 (UNIT: %)

| Algorithms | Accuracy | Precision | Recall | F1-score |
|---|---|---|---|---|
| ResNet50 | 85.83±3.73†* | 87.27±3.34†* | 85.83±3.73†* | 85.65±3.76†* |
| FedAvg | 90.84±1.86†* | 92.15±2.81†* | 90.84±1.86†* | 90.60±1.97†* |
| FedProx | 91.67±2.95†* | 93.39±2.24†* | 91.67±2.95†* | 91.41±3.03†* |
| FedBN | 92.50±3.48†* | 93.52±3.75†* | 92.50±3.48†* | 92.31±3.65†* |
| MOON | 92.50±1.86†* | 94.11±1.35†* | 92.50±1.86†* | 92.25±1.99†* |
| **FL-BT** | **94.17±1.86*** | **95.36±1.47*** | **94.17±2.28*** | **94.05±2.39*** |
| **SSL-FL-BT** | **96.66±1.86** | **97.14±1.60** | **96.66±1.86** | **96.64±1.88** |

TABLE IV
CLASSIFICATION RESULTS OF DIFFERENT ALGORITHMS ON C3 (UNIT: %)

| Algorithms | Accuracy | Precision | Recall | F1-score |
|---|---|---|---|---|
| ResNet50 | 89.75±2.05†* | 90.45±1.88†* | 89.75±2.05†* | 89.74±2.14†* |
| FedAvg | 90.00±2.65†* | 91.11±2.46†* | 90.00±2.65†* | 90.00±2.58†* |
| FedProx | 90.25±2.05†* | 90.98±1.82†* | 90.25±2.05†* | 90.18±2.08†* |
| FedBN | 90.75±2.88†* | 91.54±2.50†* | 90.75±2.88†* | 90.78±2.77†* |
| MOON | 91.25±1.53†* | 91.21±1.37†* | 91.25±1.53†* | 91.25±1.41†* |
| **FL-BT** | **92.50±0.88*** | **93.02±1.11*** | **92.50±0.88*** | **92.47±0.88*** |
| **SSL-FL-BT** | **94.50±0.68** | **94.85±0.83** | **94.50±0.68** | **94.46±0.69** |

Noting: the † and * denote the improvements achieved by FL-BT and SSL-FL-BT, respectively, are statistically significant.

Fig. 6 shows the PR curves and the corresponding AP values for different algorithms. The proposed SSL-FL-BT achieves the best AP value of 0.9864 on C1, 0.9703 on C2, and 0.9591 on C3, which again indicates its effectiveness.

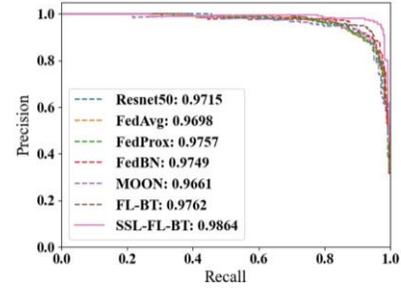

(a) C1

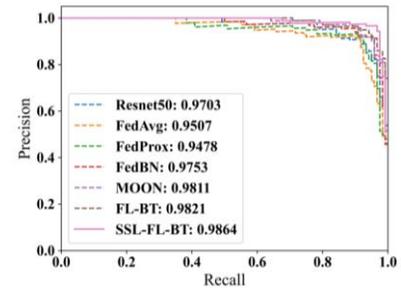

(b) C2

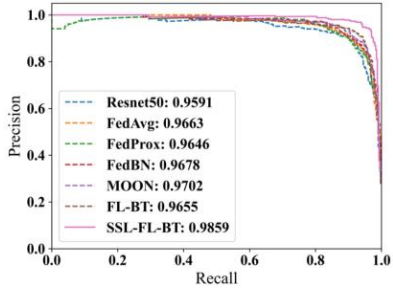

(c) C3

Fig. 6. PR curves of the compared algorithms with the corresponding AP values on the datasets of (a) C1, (b) C2 and (c) C3.

### B. Results of Ablation Experiments

Table V to VII show the results of ablation experiments on C1, C2 and C3 datasets, respectively. These experiments evaluate the effectiveness of the performance of multi-task SSL in SSL-FL-BT compared to SSL-C-FL-BT and SSL-R-FL-BT. Both the SSL-C-FL-BT and SSL-R-FL-BT improve their performances compared to FL-BT, indicating that the single pretext task, namely center classification task or image restoration task, effectively promotes the feature representation of backbone for final classification task. Moreover, SSL-C-FL-BT achieves a little better improvement compared to SSL-R-FL-BT, suggesting that the specific and heterogeneous information provided by the center-source classification task can further assist the diagnosis. Besides, our proposed SSL-FL-BT improvs at least 1.65%, 1.22%, 1.62%, and 1.61% on the accuracy, precision, recall, and F1-score, respectively, on C1 dataset, 1.66%, 1.78%, 1.66%, and 1.72% on C2 dataset, respectively, on C2 dataset, and 0.75%, 0.55%, 0.75%, and 0.77%, respectively, on C3 dataset on the corresponding indices over other compared algorithms. It demonstrates the effectiveness of the multi-task SSL framework in FL.

Tables VIII to X show the results of SSL-FL compared to other SSL algorithms, including SSL-Fed-R, SSL-Fed-S, FedSSL-R, and FedSSL-R. Table VIII gives the results of ablation experiments on C1. SSL-FL improves of 1.06%, 0.68%, 1.06%, and 1.08% on accuracy, precision, recall, and F1-score. In Table IX, SSL-FL promotes the classification accuracy, precision, recall, and F1-score by at least 1.67%, 0.89%, 1.67%, and 1.69% on C2, respectively. Moreover, SSL-FL gets at least 1.00%, 0.93%, 1.00%, and 1.01% improvements over FedSSL-S on C3.

TABLE V
ABLATION EXPERIMENT RESULTS ON C1 (UNIT: %)

| Algorithms | Accuracy | Precision | Recall | F1-score |
|---|---|---|---|---|
| FL-BT | 93.42±1.33* | 94.04±1.52* | 93.64±1.22* | 93.47±1.46* |
| SSL-C-FL-BT | 94.41±1.55* | 94.99±1.32* | 94.53±1.22* | 94.42±1.58* |
| SSL-R-FL-BT | 94.16±2.29* | 94.75±2.19* | 94.36±2.18* | 94.20±2.36* |
| **SSL-FL-BT** | **96.06±0.57** | **96.21±0.46** | **96.15±0.53** | **96.03±0.56** |

TABLE VI
ABLATION EXPERIMENT RESULTS ON C2 (UNIT: %)

| Algorithms | Accuracy | Precision | Recall | F1-score |
|---|---|---|---|---|
| FL-BT | 94.17±1.86* | 95.36±1.47* | 94.17±2.28* | 94.05±2.39* |
| SSL-C-FL-BT | 95.00±1.86* | 95.36±1.20* | 95.00±1.86* | 94.92±1.95* |
| SSL-R-FL-BT | 95.00±3.48* | 96.07±2.57* | 95.00±3.48* | 94.89±3.60* |
| **SSL-FL-BT** | **96.66±1.86** | **97.14±1.60** | **96.66±1.86** | **96.64±1.88** |

TABLE VII
ABLATION EXPERIMENT RESULTS ON C3 (UNIT: %)

| Algorithms | Accuracy | Precision | Recall | F1-score |
|---|---|---|---|---|
| FL-BT | 92.50±0.88* | 93.02±1.11* | 92.50±0.88* | 92.47±0.88* |
| SSL-C-FL-BT | 93.75±1.53* | 94.30±1.37* | 93.75±1.53* | 93.69±1.49* |
| SSL-R-FL-BT | 93.50±1.85* | 93.91±1.94* | 93.50±1.85* | 93.42±1.84* |
| **SSL-FL-BT** | **94.50±0.68** | **94.85±0.83** | **94.50±0.68** | **94.46±0.69** |

TABLE VIII
ABLATION EXPERIMENT RESULTS ON C1 (UNIT: %)

| Algorithms | Accuracy | Precision | Recall | F1-score |
|---|---|---|---|---|
| ResNet50 | 89.26±2.33†* | 90.31±2.68†* | 89.21±2.08†* | 89.26±2.24†* |
| FedAvg | 90.83±1.50†* | 91.84±1.79†* | 90.83±1.56†* | 90.73±1.65†* |
| SSL-Fed-R | 91.84±1.42†* | 92.48±1.55†* | 92.08±1.41†* | 91.92±1.56†* |
| SSL-Fed-S | 92.58±1.26†* | 93.20±1.11†* | 92.83±1.19†* | 92.65±1.41†* |
| FedSSL-R | 93.55±2.29†* | 94.20±2.13†* | 93.75±2.19†* | 93.59±2.42†* |
| FedSSL-S | 93.94±2.04†* | 94.70±1.70†* | 94.03±2.03†* | 93.99±2.18†* |
| **SSL-FL** | **94.94±0.98*** | **95.38±0.98*** | **95.09±1.02*** | **95.07±1.01*** |
| **SSL-FL-BT** | **96.06±0.57** | **96.21±0.46** | **96.15±0.53** | **96.03±0.56** |

TABLE IX
ABLATION EXPERIMENT RESULTS ON C2 (UNIT: %)

| Algorithms | Accuracy | Precision | Recall | F1-score |
|---|---|---|---|---|
| ResNet50 | 85.83±3.73†* | 87.27±3.34†* | 85.83±3.73†* | 85.65±3.76†* |
| FedAvg | 90.84±1.86†* | 92.15±2.81†* | 90.84±1.86†* | 90.60±1.97†* |
| SSL-Fed-R | 91.67±2.95†* | 92.99±3.41†* | 91.67±2.95†* | 91.44±3.08†* |
| SSL-Fed-S | 92.50±3.48†* | 93.52±3.75†* | 92.50±3.48†* | 92.31±3.65†* |
| FedSSL-R | 92.50±1.86†* | 94.29±1.20†* | 92.50±1.86†* | 92.30±1.95†* |
| FedSSL-S | 93.33±2.28†* | 94.82±1.47†* | 93.33±2.28†* | 93.18±2.39†* |
| **SSL-FL** | **95.00±1.86*** | **95.71±1.61*** | **95.00±1.86*** | **94.87±2.08*** |
| **SSL-FL-BT** | **96.66±1.86** | **97.14±1.60** | **96.66±1.86** | **96.64±1.88** |

TABLE X
ABLATION EXPERIMENT RESULTS ON C3 (UNIT: %)

| Algorithms | Accuracy | Precision | Recall | F1-score |
|---|---|---|---|---|
| ResNet50 | 89.75±2.05†* | 90.45±1.88†* | 89.75±2.05†* | 89.74±2.14†* |
| FedAvg | 90.00±2.65†* | 91.11±2.46†* | 90.00±2.65†* | 90.00±2.58†* |
| SSL-Fed-R | 91.00±1.63†* | 91.72±1.50†* | 91.00±1.63†* | 90.98±1.72†* |
| SSL-Fed-S | 91.75±1.43†* | 92.27±1.41†* | 91.75±1.43†* | 91.74±1.35†* |
| FedSSL-R | 92.50±0.88†* | 93.03±1.09†* | 92.50±0.88†* | 92.48±0.87†* |
| FedSSL-S | 92.75±1.05†* | 93.25±1.23†* | 92.75±1.05†* | 92.70±1.01†* |
| **SSL-FL** | **93.75±0.88*** | **94.18±0.87*** | **93.75±0.88*** | **93.71±0.91*** |
| **SSL-FL-BT** | **94.50±0.68** | **94.85±0.83** | **94.50±0.68** | **94.46±0.69** |

Noting: the * denotes that SSL-FL-BT gets statistically significant improvement on this result and the † denotes that SSL-FL gets statistically significant improvement on this result.

### C. Generalization and Robustness Analysis

Fig. 7 shows the GTA results of different algorithms. It can be found that all the FL algorithms achieve superior GTA values to ResNet50, suggesting that the FL strategy can effectively improve the generalization of CAD models with multi-center data. Moreover, the proposed SSL-FL-BT algorithm achieves the best GTA values with the mean accuracy of 95.74±1.43%, precision of 96.06±0.96%, recall of 95.77±1.02%, and F1-score of 95.71±1.44%. It improves at least 3.75%, 3.33%, 3.69%, and 3.78% on the corresponding indices, respectively, over FedAvg, FedProx, FedBN, and MOON. Moreover, as shown in Fig. 7, the proposed SSL-FL-BT achieves the lowest standard deviation on all indices, indicating SSL-FL-BT has better robustness and generalization.

x

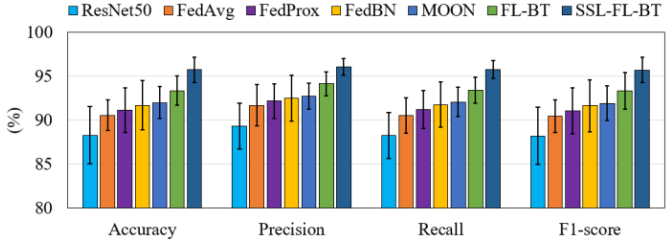

Fig. 7. Histogram chart of the GTA for the compared algorithms.

We performed additional generalization experiments, which trained the global model on C1, C2 and C3, and then fine-tuned the backbone using the Databiox Dataset on AC for generalization verification.

In TABLE XI, our proposed SSL-FL-BT achieves the best results by improving at least 1.27%, 0.78%, 1.10%, and 1.23% on the accuracy, precision, recall, and F1-score, respectively.

TABLE XI
CLASSIFICATION RESULTS OF DIFFERENT ALGORITHMS ON AC (UNIT: %)

| Algorithms | Accuracy | Precision | Recall | F1-score |
|---|---|---|---|---|
| ResNet50 | 77.68±3.36†* | 78.11±3.28†* | 78.03±3.28†* | 77.56±3.58†* |
| FedAvg | 78.00±3.21†* | 78.15±3.20†* | 78.03±3.21†* | 77.86±3.22†* |
| FedProx | 78.62±3.48†* | 79.26±3.34†* | 78.68±4.18†* | 78.52±3.86†* |
| FedBN | 78.77±2.66†* | 78.94±3.21†* | 78.92±2.81†* | 78.76±2.78†* |
| MOON | 79.29±2.95†* | 79.08±3.62†* | 78.75±2.63†* | 78.52±3.04†* |
| **FL-BT** | **80.21±2.59*** | **80.86±3.02*** | **80.42±2.46*** | **80.22±2.48*** |
| **SSL-FL-BT** | *81.48±2.51* | *81.64±2.41* | *81.52±2.62* | *81.45±2.53* |

## VI. DISCUSSION

In this work, a novel SSL-FL-BT framework is proposed to promote both the diagnostic accuracy and generalization ability of the CAD model for histopathological images. The experimental results on three public datasets have validated the effectiveness of the proposed SSL-FL-BT.

Existing FL paradigm only shares the model parameters of different centers to improve the model performances. However, it cannot guarantee that the distributed CAD models well capture the specific properties of data from different centers. Therefore, we break through this limitation by sharing not only the model parameters in the central server, but also the pseudo histopathological images generated from each center, because they contain inherent and specific properties corresponding to the real images in this center, but do not include the privacy information. We therefore consider it the most important contribution in this work.

The stain of histopathological images is generally affected by several factors, such as the brand, storage time, and temperature, which result in the inconsistencies of histopathological images across different hospitals. It then degrades the generalization ability of the CAD model, if this model is only trained with the data acquired from a single center. On the other hand, it is generally time-consuming and expensive to collect large amounts of annotated data in one center, and therefore, the SSS problem is common in the field of CAD. Multi-center learning is an effective way to alleviate both issues, which trains the model by collecting data from different centers. It then not only enriches the variety of histopathological images, but also enhances the training set. Therefore, the multi-center learning based CAD is more feasible to meet the clinical requirement than the single-center based approach, and it can effectively improve both the diagnostic accuracy and generalization ability. In this work, the experimental results in Table II to IV demonstrate that FL algorithms achieve superior performances to the CAD model trained only using single-center data, and also protect the private information.

Although FL has the advantage of privacy protection, it still cannot guarantee that the distributed CAD models fully capture the center-specific information from each center's data to further improve generalization, because it does not share the data from different centers. In this work, we propose to generate pseudo histopathological images from different centers for this issue. These self-generated pseudo images contain center-specific inherent properties corresponding to the original images of each center, respectively, but do not have private information. Therefore, these pseudo images can be shared in the central server. Two SSL pretext tasks are then designed based on the characteristics of pseudo images for the pre-training backbone. The center classification task tries to learn the heterogeneous properties of each center, and the image restoration task aims to capture common characteristics across centers. The results in Table V, Table VI, and Table VII show that the classification task can better promote the performance compared to the restoration task. It seems that the center-specific information is more helpful for the generalization of a CAD model. Besides, the combination of two tasks achieves significant improvement compared with the single task, demonstrating the effectiveness of multi-task SSL driven by the properties of multi-center data themselves.

As shown in Table VIII to X, both the FedSSL-R and FedSSL-S algorithms achieve superior performance over the corresponding SSL-Fed-R and SSL-Fed-S. In the first stage, the FedSSL-based variants conduct the SSL in the FL framework, and thus well train the initial models of different centers for the following FL, while the SSL-Fed-based variants only implement SSL once to initialize the model of each center for the following FL. Therefore, the FedSSL-based approaches can learn more information from other centers than the SSL-Fed-based variants to initialize the model. Moreover, although the proposed SSL-FL and SSL-FL-BT also only implement the SSL once, they still outperform both the FedSSL- and SSL-Fed-based variants, because of the following two reasons: 1) the SSL is directly conducted in the central server on all the pseudo images from different centers; and 2) the proposed multi-task SSL further promotes the model to learn the common representations and specific properties.

In addition, we propose an effective algorithm, namely FL-BT, to improve the classification performance of local training. FL-BT utilizes the similarity between model representations to minimize the representation redundancy of the local model, which benefits the optimization of the global model in the FL procedure. It is known that both insufficient and excessive negative sample pairs will affect the performance of contrastive learning [31]. Our FL-BT avoids these problems, since no negative samples are required in BT algorithm. Moreover, the learning criterion of FL-BT tries to obtain a feature representation that contains more information, so that the dimension of each feature preferably has an independent meaning. On the other hand, as shown in Eq. (5), the loss function in FL-BT consists of two parts, *i.e.*, the invariance term and redundancy reduction term. The former plays a role in bringing the positive examples closer to each other in the representation space, while the latter enhances the independence of each element of the vector. Compared with the conventional contrastive SSL algorithms, which require positive and negative samples to conduct contrastive learning, the proposed FL-BT eliminates the redundant information expression in the representation vector as much as possible. Therefore, FL-BT achieves superior performance.

In the proposed SSL-FL-BT, SSL and FL-BT are two independent operations for different purposes, but both effectively improve the generalization of global model. Specifically, SSL provides an initialization global model for the subsequent FL stage, which improves model generalization from the perspective of a pre-trained model with pseudo from different centers. FL-BT maximizes the consistency between the representation learned by the local model and the representation learned by the global model to correct the local training, thereby improving the generalization of the global model.

Although the results on three public datasets indicate the superior performance of the proposed SSL-FL-BT framework, it still has room for improvement. For example, the proposed framework performs the patch-level diagnosis for histopathological images in this work, while the WSI-based CAD has attracted considerable attention in recent years, which is more difficult due to the SSS problem. In fact, MIL is a commonly used method for the classification task of WSIs. Specifically, each WSI is regarded as a bag and the numerous cropped patches in this WSI are used as instances. Therefore, MIL is also the patch-based method for WSIs [48]. Consequently, it is also feasible to further extend the proposed framework to the MIL-based CAD for WSIs, which is our future work. On the other hand, the current pseudo-data based SSL is performed on the central server, and it can be further improved with the SSL training manner by combining both the central server and distributed centers.

## VII. CONCLUSION

In this work, a novel pseudo-data based SSL-FL-BT framework is proposed to improve both the diagnostic accuracy and generalization of the CAD model for histopathological images. The self-generated pseudo images contain inherent and center-specific properties corresponding to the real histopathological images of each center without privacy information, while the self-designed multi-task SSL captures both the representation from these pseudo images for the pre-trained backbone network. The experimental results on three public histopathological image datasets indicate the effectiveness of the proposed SSL-FL-BT.